\documentclass[conference]{IEEEtran}
\IEEEoverridecommandlockouts
\usepackage{cite}
\usepackage{amsmath,amssymb,amsfonts}
\usepackage{algorithmic}
\usepackage{graphicx}
\usepackage{textcomp}
\usepackage{xcolor}
\usepackage{url}

\usepackage{mathtools} 
\usepackage{bm}        
\usepackage{tabularx}
\usepackage{array}    
\usepackage{lipsum}   
\newcolumntype{Y}{>{\centering\arraybackslash}X}
\newcolumntype{C}{>{\centering\arraybackslash}m{\dimexpr.5\linewidth-2\tabcolsep}}
\def\BibTeX{{\rm B\kern-.05em{\sc i\kern-.025em b}\kern-.08em
    T\kern-.1667em\lower.7ex\hbox{E}\kern-.125emX}}
\begin{document}

\title{Introducing cosmosGPT: Monolingual Training for Turkish Language Models
}


\author{
    \IEEEauthorblockN{
        H. Toprak Kesgin\IEEEauthorrefmark{1},
        M. Kaan Yuce\IEEEauthorrefmark{1},
        Eren Dogan\IEEEauthorrefmark{1},
        M. Egemen Uzun\IEEEauthorrefmark{1},
        Atahan Uz\IEEEauthorrefmark{1}\\
        H. Emre Seyrek\IEEEauthorrefmark{1},
        Ahmed Zeer\IEEEauthorrefmark{1},
        M. Fatih Amasyali\IEEEauthorrefmark{1}
    }
    \IEEEauthorblockA{\IEEEauthorrefmark{1}Cosmos AI Research Group, Department of Computer Engineering, Yildiz Technical University, Istanbul, Turkey}
}

\maketitle

\begin{abstract}
The number of open source language models that can produce Turkish is increasing day by day, as in other languages. In order to create the basic versions of such models, the training of multilingual models is usually continued with Turkish corpora. The alternative is to train the model with only Turkish corpora. In this study, we first introduce the cosmosGPT models that we created with this alternative method. Then, we introduce new finetune datasets for basic language models to fulfill user requests and new evaluation datasets for measuring the capabilities of Turkish language models. Finally, a comprehensive comparison of the adapted Turkish language models on different capabilities is presented. The results show that the language models we built with the monolingual corpus have promising performance despite being about 10 times smaller than the others.
\end{abstract}


\section{Introduction}
Recent advances in artificial intelligence and natural language processing (NLP) have underscored the effectiveness of language models trained on extensive datasets. 
However, the majority of these studies have centered on languages with rich bibliographies, such as English. 
The success observed in large English-based language models suggests that similar potential might be unlocked for other languages. 
This study examines cosmosGPT Medium and cosmosGPT Large models trained exclusively on Turkish, a relatively low-resource language. 
Additionally, this research introduces new datasets and evaluation metrics specifically designed to address the distinctive needs of such languages. 
Other models like TURNA\cite{uludougan2024turna}, Kanarya\cite{safaya-etal-2022-mukayese}, and VBart\cite{turker2024vbart} continue to contribute to natural language understanding and production in Turkish. 
However, there are no instruction-following versions of these models.

The existing literature on language models trained from scratch for Turkish indicates that there is still considerable potential for improvement in this area. 
However, the models that have been developed to date have been few in number and have not yet demonstrated significant success. 
As a result of these limitations, researchers have increasingly turned to more comprehensive and large-scale multilingual models. 
However, these models require larger structures and often perform less well than monolingual models in terms of language-specific tasks. 
For instance, the mGPT\cite{shliazhko2024mgpt}, despite its impressive scope covering 61 languages, often falls short in capturing the nuances of each language compared to dedicated monolingual models. 
Due to the fact that multilingual models are trained for a multitude of languages and their tokenizers are frequently inefficient, this creates significant disadvantages for languages such as Turkish\cite{turker2024vbart}. These issues impede progress in the field of NLP and render such models both costly and challenging to utilise in practice.

Carefully choosing the model architecture and preparing the training data are critical aspects that shape both the quality of outcomes and the performance of the language model. 
The dataset used affects the model’s generalization and inference capabilities, as well as the efficacy of the learning process. 
In particular, for large language models, extensive, diverse, and high-quality datasets are essential to ensure comprehensive coverage of varied linguistic structures, facilitating a broader understanding of language.
In this context, two GPT-2 base models \cite{radford2019language} were trained, mirroring the openAI GPT-2 configurations with 355 million and 774 million parameters, respectively. 
These models were trained using a 250GB dataset derived from web-based CulturaX \cite{nguyen2023culturax} sources and a 25GB dataset compiled from books and other internet resources. 

The advent of models such as ChatGPT\cite{achiam2023gpt}, Llama\cite{touvron2023llama}, and Mistral \cite{jiang2023mistral} has led to the prominence of the instruction completion chatbot features of such GPT like models. 
In this context, we created instruct versions of the models we trained and compared their capabilities with much larger multi-language Turkish finetuned versions, using human voting and automated metrics. 
Our findings demonstrated that these models were competitive with models with 10 times more parameters.

This research tackles the challenges associated with training large language models, particularly focusing on Turkish language. 
The GPT-2 models we developed have shown a marked improvement over existing multilingual models, particularly in capturing the intricacies and nuances of the Turkish language. 
These models have proven versatile, finding applications in language learning, automatic text generation, and semantic analysis, thus broadening their practical utility. 
Notably, the specialized datasets utilized in our fine-tuning processes have significantly bolstered the models’ ability to comprehend and execute instructions. 
Moreover, the data collected in our evaluation phase allowed us to measure the effectiveness of the models.

Our study aims to fill important gaps and make significant in the field of NLP in Turkish. The main contributions of our research are outlined below:
\begin{enumerate}
        \item Dedicated Turkish-only cosmosGPT Medium and cosmosGPT Large models have been developed from scratch, including their instruction-completion versions, and have been made available as open source.
        \item New fine-tuning and evaluation datasets have been developed to enhance the models' adaptability to various instruction execution tasks in Turkish and to objectively assess texts. Extensive analyses on these datasets have further refined our understanding and performance measurement of the models.
        \item A comprehensive comparison of existing large language models available for Turkish has been conducted, highlighting the differences and strengths/weaknesses among these models. It has been demonstrated that our models trained specifically for Turkish can outperform models with up to ten times more parameters.
        \item Correlations between human evaluations and other criteria have been analyzed during the model evaluation process.
\end{enumerate}

The combination of these contributions has resulted in a notable advancement in the field of Turkish NLP, offering researchers the chance to develop more effective models.  

\section{Training GPT models in Turkish}
The development of large language models relies heavily on extensive and detailed training datasets and large model sizes. 
The success of these models is directly linked to the quality and size of the datasets used.
Particularly in multilingual scenarios, the available multilingual text datasets may not be sufficiently cleaned or organized. 
This issue leads to a lack of transparency in the datasets used for training language models, complicating progress in research.

For training our Turkish language model, we opted for the meticulously cleaned and deduplicated CulturaX dataset, in contrast to the often noisy data derived from general web sources. This dataset provides a large corpus cleansed of undesirable content, unlike the noisy data collected from the web. CulturaX has undergone a multi-stage cleaning process including language recognition, URL-based filtering, metric-based cleaning, document refinement, and data deduplication. These processes aim to provide the highest quality data for model training. We further refined CulturaX by filtering out slang, eventually compiling a dataset of 250GB. We then added our own compiled 25GB dataset from books, forums, news sites, and other sources, reaching a total training corpus of 275GB. This comprehensive 275GB training corpus offers a rich and diverse array of linguistic structures, enabling the model to cover a broader spectrum of language. We trained our tokenizer on 100GB of this data with a vocabulary size of 50,257.

In our research, we utilised the GPT-2 model architecture developed by OpenAI. The cosmosGPT Medium model, configured with 355 million parameters, includes 24 layers, 16 attention heads, and an embedding size of 1024. In contrast, the cosmosGPT Large model, with 774 million parameters, comprises 36 layers, 20 attention heads, and an embedding size of 1280. Both models are designed with a context length of 1024 tokens, utilise the GELU new activation function, and are structured to handle a comprehensive range of linguistic details.

The training of the models was conducted using Google Cloud's TPUv3-8 infrastructure. The Large model was trained with a batch size of 16 and for 2 epochs, while the Medium model used a batch size of 128 and was trained for 3 epochs. The training process was optimized using an Adam optimizer with a learning rate of $1e-4$ with linear decay. This approach has enabled both rapid and efficient training of the models.

In conclusion, this training process and the use of high-quality datasets have enabled us to achieve significant advancements in the accuracy and reliability of our Turkish language models.

\section{Creating Finetune Dataset}
To create instruction versions of the models and develop the best finetuning datasets, we have compiled various datasets and merged shared collections.
Finding a suitable dataset for finetuning has been one of the focal points of our research. Original datasets have been created using various sources, including open-source platforms, contributions from individuals, and data generated by GPT-4\cite{achiam2023gpt}. Initially, four primary datasets were established. Subsequently, through cleaning and various combinations, new datasets have been derived. Here are the names and features of the mentioned primary datasets:

\vspace{0.25\baselineskip}

\small{

    \begin{itemize}
    
    \item $\mathbf{Merve/turkish\_instructions \; ( M ) :}$ An open-source dataset comprising various instructions, consisting of 51,000 unique entries\cite{merveturkishinstructions} translated from alpaca dataset \cite{alpaca}.
    
    \vspace{0.5\baselineskip}
    
    \item $\mathbf{BactrianX \; ( B ) :}$ A dataset obtained by translating 67K English instructions (alpaca-52k + dolly-15k) into 51 languages using the Google Translate API \cite{li2023bactrianx}. The responses to these instructions were generated by GPT-3.5.
    
    \vspace{0.5\baselineskip}
    
    \item $\mathbf{Human \; ( H ) :}$ A dataset with original instructions written by a human. Both human and machine responses (from ChatGPT and Gemini \cite{team2023gemini}) have been recorded, totaling 16,000 entries.
    
    \vspace{0.5\baselineskip}
    
    \item $\mathbf{GPT4 \; ( g ) :}$ CustomGPT was created using fixed prompts and hyperparameters. For this CustomGPT, 5,000 entries from the M dataset were generated.
    
    \end{itemize}
}
\vspace{0.25\baselineskip}

From these primary datasets, \textit{filtered} datasets and various \textit{combinations} have been derived.

\subsection{Filtering \& Combination}
Discussing the filtering method, a BERT\cite{toprak2023developing} model was fine-tuned using data handpicked as high-quality by human reviewers for classification purposes. This model was then employed to sift through the larger dataset, discarding entries that fell below a certain threshold. Consequently, $\mathbf{M}$ dataset was refined into $\prescript{\bm{f}}{}{\bm{M}}$ (Filtered $\mathbf{M}$). Similarly, $\mathbf{B}$ dataset has been refined to $\prescript{\bm{f}}{}{\bm{B}}$ (Filtered $\mathbf{B}$). \newline

To observe the effects of diversity, different combinations were subsequently formed: \newline

\small{
 $ \mathbf{ 
    B \; + \; M \; \rightarrow  \; BM     \; \; \; \; \; \;  \; \; \; \; \; \; \; \; \; \; \; \; \; \; \; \; \; \;
    M \; + \; H \; \rightarrow  \; MH
    }
 $
} 

\vspace{0.5\baselineskip}

\small{
 $ \mathbf{ 
    \prescript{\bm{f}}{}{\bm{B}} \; + \; H \; + \; M \; \rightarrow \; \prescript{\bm{f}}{}{\bm{B}}HM 
    \; \; \; \; \; \;
    \prescript{\bm{f}}{}{\bm{B}} \; + \; \prescript{\bm{f}}{}{\bm{M}} \; \rightarrow \; f^2 BM
    }
 $
}
\vspace{0.5\baselineskip}

\small{
 $ \mathbf{ 
        g \; + \; B \; + \; H + \; M \rightarrow\; gBHM
    }
 $
} \newline

\section{Creating Evaluation Datasets}

To measure the performance of models trained with the finetuning dataset, two test sets have been created. The test set used for human voting has been divided into various categories and consists of 400 questions. Let's refer to this dataset as $\bm{V}$ (Voting). The other dataset is more general and consists of 1000 instructions. Let's refer to it as $\bm{G}$ (General). Divided into 13 categories, each category contains an average of 30 instructions. 
Some categories are exemplified in Table ~\ref{Vtestkumesitablosu}.

\newcolumntype{K}{>{\centering\arraybackslash}X}

\begin{table}[htbp]
\caption{V Test Dataset}
\begin{center}
\begin{tabular}{|C|C|}
\hline
\textbf{Category} & \textbf{Instruction} \\
\hline
Hikaye Oluşturma & Engelli bir genç profesyonel bir atlet olmak ister. Çektiği zorlukları anlat. \\
\hline
Benzerlik Bulma & Aşağıdaki listede çorap, hangilerine uymamaktadır? Bisiklet, Gömlek, Tren, Kitap, Uçak. \\
\hline
Başlık Oluşturma & Bir kahve dükkanında geçen romantik bir hikaye. Bu hikaye için çekici bir başlık oluşturun. \\
\hline
... & ... \\
\hline
Basit Matematik & 100 gramı 5 TL olan fındığın kilosu kaç TL'dir? \\
\hline
\end{tabular}
\label{Vtestkumesitablosu}
\end{center}
\end{table}

\begin{table}[htbp]
\caption{G Test Dataset}
\begin{center}
\begin{tabular}{|C|C|}
\hline
\textbf{Instruction} & \textbf{Reference Answer} \\
\hline
Türkiye’nin başkenti neresidir? & Türkiye’nin baskenti Ankara’dır \\
\hline
Aristotales ve Platon arasındaki
ilişki nedir? & Aristotales Platon’un öğrencisidir. \\
\hline
... & ... \\
\hline
Yapay zeka işsizlik riski yaratıyor mu? & Evet, yapay zeka teknolojileri tekrar eden işleri yapan çalışanları işsiz bırakabilir. \\
\hline
\end{tabular}
\label{your_label_here}
\end{center}
\end{table}


\subsection{Evaluation Criteria}

In this study, the metrics and human evaluation methods used to assess the performance of Turkish language models have been thoroughly examined. These selected metrics and methods are employed to elucidate and compare performance differences among language models in an objective and comparative manner. For this purpose, automatic metrics such as the ROUGE-1, ROUGE-2, ROUGE-L and cosine similarity metrics have been calculated between the actual answers and the model responses. The vector representations for cosine similarity were computed using the all-MiniLM-L12-v2 model \cite{sentencetransformers}.

In human evaluations, metrics such as the ELO score and Win Percentage were utilized to further assess model performance. ELO score is a metric used to evaluate the performance of language models against each other based on human voting. During the voting process, each judge evaluated the responses given by the models to questions in the $\mathbf{V}$ dataset. Each step randomly presented a question and two different model responses to the judge. To ensure judges remained objective, the names of the models were concealed. The ELO scoring began with each model rated at 1000. In matchups, the preferred model gained ELO points while the other lost points. Winning against a model with a high ELO score yields more points, whereas winning against a model with a low ELO score yields fewer points. The ELO system is particularly effective in understanding the balance of power between models. A high ELO score indicates that a model performs better compared to others. To better observe ELO results, the order of matchups in the current dataset was randomly rearranged, and the ELO scores of models were recalculated for 1000 different scenarios. Each permutation represents a scenario where matchups are conducted in a completely random order. Confidence intervals and averages for each model's ELO values were calculated across these 1000 different scenarios, thereby broadening the investigation of potential impacts of matchup orders on ELO scores and enhancing the generalizability of the results. The ELO ratings were conducted through equal participation by eight different judges, with a total of $X$ votes cast, ensuring a balanced and comprehensive evaluation. Win Percentage (Winpct) is a ratio that shows how successful a model is against other models based on human voting. This metric calculates the ratio of votes a model receives to the total number of votes:

\begin{center}
\normalsize{$\text{{winpct}} = \frac{{\text{{win}} + \text{{both}}}}{{\text{{total}}}}$}
\end{center}

These evaluation metrics and methods allow us to objectively compare how language models perform across various scenarios and tasks.

\section{Comparison of Fine-tune Datasets Over GPT-2 Large Models}

In order to effectively determine the most suitable fine-tune dataset, different GPT-2 Large models were obtained by fine-tuning each dataset created. The fine-tuning process was carried out for 3 epochs for each model. For this purpose, a total of ten distinct models have been developed. The responses of these models to the G and V evaluation datasets were measured using Cosine Similarity and ROUGE scores. Additionally, the responses given by the models to the V dataset were presented to judges for voting, and ELO and WinPct values were calculated.

In the tables and figures, each model is identified by the name of the fine-tune dataset it was trained on. The scores for the V dataset are shown in Table ~\ref{Vskorlar}, and the scores for the G dataset are presented in Table ~\ref{Gskorlar}. 

\newcolumntype{S}{>{\centering\arraybackslash\scriptsize\bfseries}m{0.1\linewidth}} 
\newcolumntype{a}{>{\centering\arraybackslash\scriptsize}m{0.05\linewidth}} 

\begin{table}[htbp]
\caption{Scores According to the 'V' Dataset}
\begin{center}
\begin{tabular}{|S|a|a|a|a|a|a|}
\hline
\textbf{Model} & \textbf{Cos} & \textbf{R-1} & \textbf{R-2} & \textbf{R-L} & \textbf{ELO} & \textbf{WP} \\
\hline
\centering{$\bm{M}$} & \centering{0.664} & 0.194 & 0.058 & 0.184 & \centering{1029} & 43.63 \\
\hline
\centering{$\prescript{\bm{f}}{}{\bm{M}}$} & 0.663 & 0.206 & 0.067 & 0.198 & \centering{979} & 41.07 \\
\hline
\centering{$\bm{B}$} & 0.631 & 0.155 & 0.043 & 0.148 & \centering{958} & 37.08 \\
\hline
\centering{$\bm{H}$} & 0.619 & 0.148 & 0.045 & 0.142 & \centering{820} & 24.08 \\
\hline
\centering{$\bm{BM}$} & 0.662 & 0.195 & 0.057 & 0.184 & \centering{1069} & 47.43 \\
\hline
\centering{$\bm{MH}$} & 0.66 & 0.187 & 0.059 & 0.179 & \centering{1071} & 44.92 \\
\hline
\centering{$\prescript{\bm{f}}{}{\bm{B}}$} & 0.644 & 0.162 & 0.052 & 0.156 & \centering{1014} & 38.57 \\
\hline
\centering{$\bm{f^2 BM}$} & 0.666 & 0.193 & 0.06 & 0.185 & \centering{1041} & 43.38 \\
\hline
\centering{$\prescript{\bm{f}}{}{\bm{BHM}}$} & 0.655 & 0.186 & 0.053 & 0.177 & \centering{1023} & 41.72 \\
\hline
\centering{ $\bm{gBHM}$ } & 0.662 & 0.183 & 0.056 & 0.176 & \centering{996} & 41.91 \\
\hline
\end{tabular}
\label{Vskorlar}
\end{center}
\end{table}

\begin{table}[htbp]
\caption{Scores According to the 'G' Dataset}
\begin{center}
\begin{tabular}{|S|a|a|a|a|}
\hline
\textbf{Model} & \textbf{Cos} & \textbf{R-1} & \textbf{R-2} & \textbf{R-L} \\
\hline
$\bm{M}$ & 0.72 & 0.205 & 0.075 & 0.195  \\
\hline
$\prescript{\bm{f}}{}{\bm{M}}$ & 0.721 & 0.211 & 0.079 & 0.2  \\
\hline
$\bm{B}$ & 0.698 & 0.175 & 0.061 &  0.167 \\
\hline
$\bm{H}$ & 0.711 & 0.189 & 0.066 & 0.179\\
\hline
$\bm{BM}$ & 0.725 & 0.208 & 0.077 & 0.197 \\
\hline
$\bm{MH}$ & 0.721 & 0.2 & 0.074 & 0.19   \\
\hline
$\prescript{\bm{f}}{}{\bm{B}}$ & 0.697 & 0.171 & 0.058 & 0.163  \\
\hline
$\bm{f^2 BM}$ & 0.722 & 0.207 & 0.077 & 0.197 \\
\hline
$\prescript{\bm{f}}{}{\bm{BHM}}$ & 0.725 & 0.201 & 0.075 & 0.191  \\
\hline
$\bm{gBHM}$ & 0.727 & 0.201 & 0.075 & 0.191\\
\hline
\end{tabular}
\label{Gskorlar}
\end{center}
\end{table}




\subsection{Score Analysis}

In this section, an analysis of the obtained results will be conducted. The first criterion of importance, due to its reliability among the evaluation criteria, will be the ELO scores resulting from judge voting. The 16,000-entry H dataset exhibited the poorest performance. However, this result is not due to the low quality of the H data set, but to the fact that the amount of data it contains is small and does not follow a harmonised format since it is collected from different people. The 51,000-entry M dataset displayed a remarkably good performance compared to the 67,000-entry B dataset, which had more data, clearly demonstrating the difference in data quality between these datasets. The B dataset was improved by filtering, but as the M dataset was already composed of high-quality data, filtering it led to the loss of good data and decreased diversity, resulting in poor outcomes. Therefore, it is not always possible to say that the filtering method used is effective in all cases. Consequently, the most suitable datasets, due to their superiority over others, have been MH and BM. Models fine-tuned with these two appropriate datasets have been selected to compete with other Turkish language models.

\section{Comparison of Turkish Language Models}

Our study has conducted an evaluation of the cosmosGPT Medium and cosmosGPT Large Turkish language models developed using the most suitable datasets identified, namely MH and BM. The goal was to analyze and compare these models against existing models in the literature that have demonstrated success in fulfilling Turkish instructions. In selecting the models for this comparison, we exclusively included those capable of responding in Turkish and which had been fine-tuned specifically for following instructions. We did not consider models that were not trained to respond to instructions. Additionally, some multilingual models that can respond in Turkish but performed very poorly when tested with our datasets were excluded from the comparison. This approach ensured a focused analysis on models most relevant to our criteria. The comparison aimed to highlight the differences between models, showcasing their relative strengths and weaknesses according to the established evaluation metrics. The models included in the comparison are those specifically trained on Turkish as well as multilingual models. The relevant models are shown in Table ~\ref{detaileddatalabel}.

\newcolumntype{C}{>{\centering\arraybackslash}m{0.15\linewidth}} 
\newcolumntype{S}{>{\centering\arraybackslash\footnotesize\bfseries}m{0.15\linewidth}} 

\setlength{\tabcolsep}{6pt} 
\begin{table}[htbp]
\caption{Detailed Table of Models}
\begin{center}
\begin{tabular}{|p{2.5cm}|C|C|C|}
\hline
\centering \textbf{Model} & \textbf{Parameter Size}  & \textbf{Base Model} & \textbf{TR Fine-Tuned} \\ \hline
\centering Turkcell-LLM-7b-v1 \cite{Trendyol} & 7.38 B & Mistral 7B\cite{jiang2023mistral} & \checkmark \\ \hline
\centering Trendyol-LLM-7b-chat-dpo-v1.0 \cite{Trendyol} & 7.34 B   & Mistral 7B &  \checkmark \\ \hline
\centering Trendyol-LLM-7b-chat-v0.1 \cite{Trendyol} & 6.84 B  & Llama 2 7B\cite{touvron2023llama2} & \checkmark \\ \hline
\centering SambaLingo-Turkish-Chat \cite{SambaLingo} & 6.95 B  & Llama 2 7B & \checkmark  \\ \hline
\centering openchat\_3.5 \cite{wang2023openchat} & 7 B  & Mistral 7B & - \\ \hline
\centering  \textbf{cosmosGPT-Medium-MH} & 355 M  & cosmosGPT Medium & \checkmark \\ \hline
\centering \textbf{cosmosGPT-Medium-BM}  & 355 M  & cosmosGPT Medium & \checkmark \\ \hline
\centering \textbf{cosmosGPT-Large-MH} & 774 M  & cosmosGPT Large & \checkmark \\ \hline
\centering \textbf{cosmosGPT-Large-BM}  & 774 M  & cosmosGPT Large & \checkmark \\ \hline
\centering gemma-7b-it \cite{googlegemma7b} & 8.54 B  & gemma-7b & - \\ \hline
\centering gemma-2b-it \cite{googlegemma2b} & 2.51 B  & gemma-2b & - \\ \hline

\end{tabular}
\label{detaileddatalabel}
\end{center}
\end{table}

\subsection{Scores of Models According to Evaluation Metrics}

\begin{figure}[htbp] 
\centerline{\includegraphics[scale=0.045]{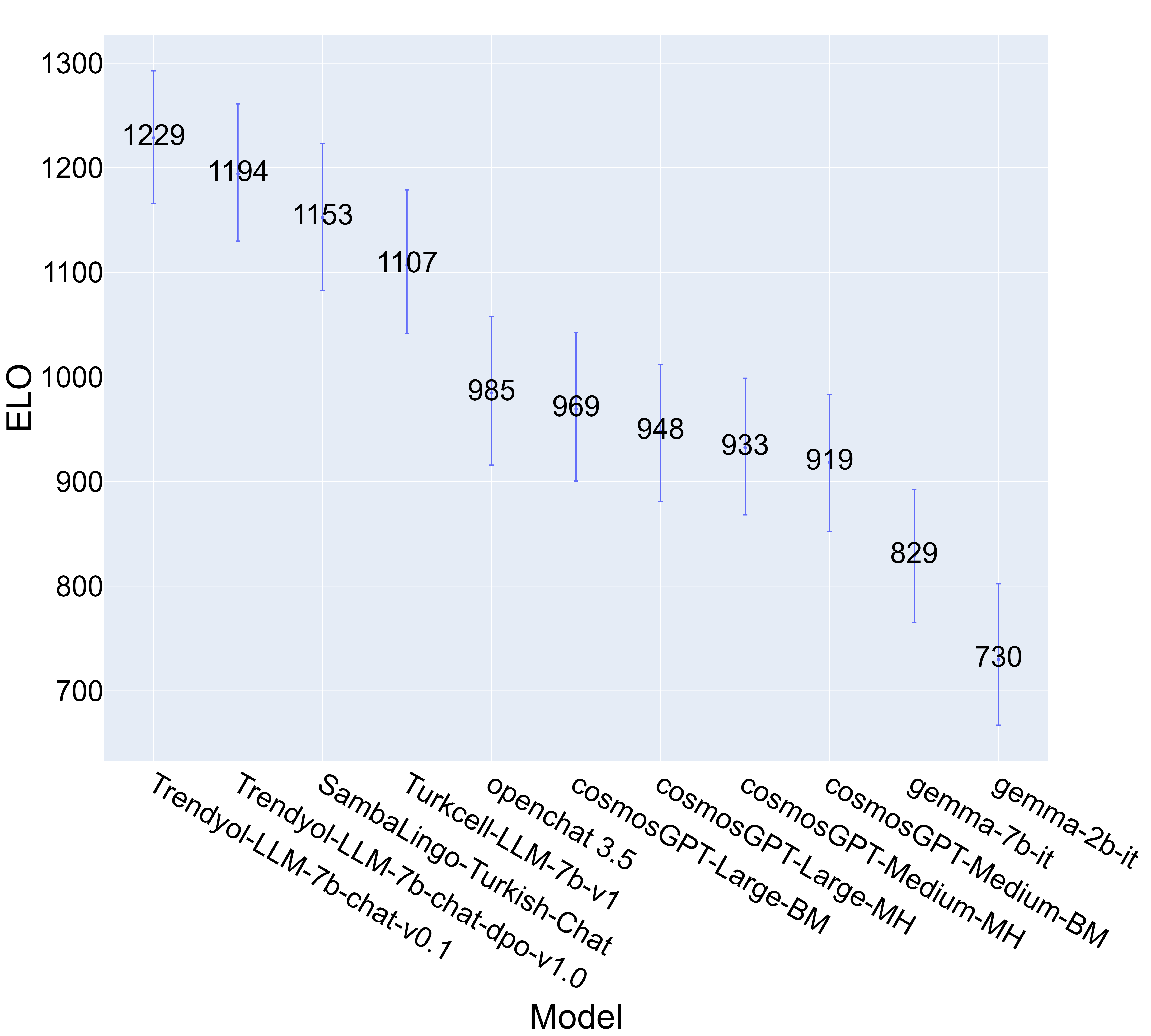}}
\caption{ELO Ratings According to The 'V' Dataset}
\label{elo_ratings_diger_modeller}
\end{figure}

The score results of each compared model according to the evaluation metrics are shown in Table~\ref{modeller22v} for dataset V and Table~\ref{1kmodeller} for dataset G. The ELO Rating results for each model with 95\% confidence interval are shown in Fig.~\ref{elo_ratings_diger_modeller}.

\begin{table}[htbp]
\caption{Scores According to the 'V' Dataset}
\begin{center}
\begin{tabular}{|p{2.5cm}|p{0.5cm}|p{0.5cm}|p{0.5cm}|p{0.5cm}|p{0.5cm}|p{0.5cm}|}
\hline
\centering \textbf{Model} & \textbf{Cos} & \textbf{R-1} & \textbf{R-2} & \textbf{R-L} & \textbf{ELO} & \textbf{WP} \\
\hline
\centering{Trendyol-LLM-7b-chat-v0.1} & 0.728 & 0.221 & 0.079 & 0.211 & \centering{1229} & 71.41 \\
\hline

\centering{Trendyol-LLM-7b-chat-dpo-v1.0} & 0.717 & 0.187 & 0.07 & 0.178 & \centering{1194} & 70.34 \\
\hline

\centering{SambaLingo-Turkish-Chat} & 0.693 & 0.185 & 0.071 & 0.179 & \centering{1163} & 65.2 \\
\hline
\centering{Turkcell-LLM-7b-v1} & \centering{0.693} & 0.191 & 0.07 & 0.177 & \centering{1107} & 57.28 \\
\hline
\centering{openchat\_3.5} & 0.685 & 0.196 & 0.072 & 0.189 & \centering{985} & 46.86 \\
\hline
\centering{cosmosGPT-Large-BM} & 0.724 & 0.208 & 0.076 & 0.197 & \centering{969} & 43.19 \\
\hline
\centering{cosmosGPT-Large-MH} & 0.721 & 0.199 & 0.073 & 0.19 & \centering{948} & 43.44 \\
\hline
\centering{cosmosGPT-Medium-MH} & 0.655 & 0.186 & 0.054 & 0.177 & \centering{933} & 41.71 \\
\hline
\centering{cosmosGPT-Medium-BM} & 0.661 & 0.197 & 0.064 & 0.186 & \centering{919} & 40.17 \\
\hline
\centering{gemma-7b-it} & 0.645 & 0.156 & 0.053 & 0.151 & \centering{829} & 28.11 \\
\hline
\centering{gemma-2b-it} & 0.621 & 0.154 & 0.051 & 0.15 & \centering{730} & 21.72 \\
\hline
\end{tabular}
\label{modeller22v}
\end{center}
\end{table}

\begin{table}[htbp]
\caption{Scores According to the 'G' Dataset}
\begin{center}
\begin{tabular}{|p{2.5cm}|p{0.5cm}|p{0.5cm}|p{0.5cm}|p{0.5cm}|}
\hline
\centering \textbf{Model} & \textbf{Cos} & \textbf{R-1} & \textbf{R-2} & \textbf{R-L} \\
\hline
\centering{Turkcell-LLM-7b-v1} & \centering{0.766} & 0.192 & 0.081 & 0.18 \\
\hline
\centering{Trendyol-LLM-7b-chat-dpo-v1.0} & 0.773 & 0.245 & 0.102 & 0.232 \\
\hline
\centering{Trendyol-LLM-7b-chat-v0.1} & 0.764 & 0.193 & 0.078 & 0.183  \\
\hline
\centering{SambaLingo-Turkish-Chat} & 0.765 & 0.199 & 0.082 & 0.189  \\
\hline
\centering{openchat\_3.5} & 0.732 & 0.20 & 0.086 & 0.193 \\
\hline
\centering{cosmosGPT-Medium-MH} & 0.721 & 0.202 & 0.077 & 0.192 \\
\hline
\centering{cosmosGPT-Medium-BM} & 0.726 & 0.214 & 0.085 & 0.203 \\
\hline
\centering{cosmosGPT-Large-MH} & 0.72 & 0.199 & 0.073 & 0.189 \\
\hline
\centering{cosmosGPT-Large-BM} & 0.724 & 0.207 & 0.076 & 0.197 \\
\hline
\centering{gemma-7b-it} & 0.699 & 0.186 & 0.08 & 0.179 \\
\hline
\centering{gemma-2b-it} & 0.675 & 0.198 & 0.08 & 0.191  \\
\hline
\end{tabular}
\label{1kmodeller}
\end{center}
\end{table}

\subsection{Categorical Task Based Comparison}

\begin{figure}[htbp] 
\centerline{\includegraphics[scale=0.4]{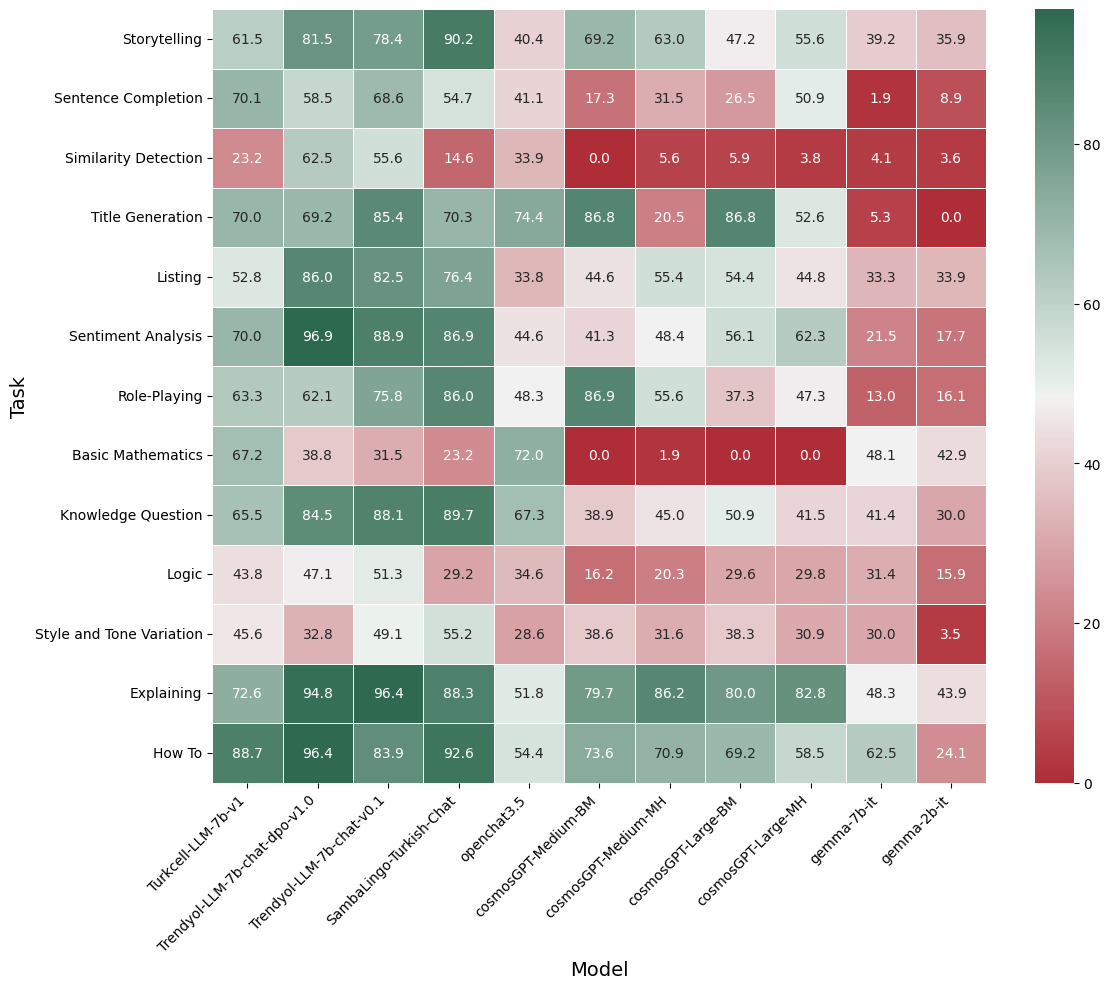}}
\caption{Scores According to Categorical Tasks}
\label{tasks_models}
\end{figure}

WinPct metric results were calculated separately for each model based on the task categories defined in the test dataset in order to measure the specific task performance of the models. This approach allows for the identification of categories where models exhibit strengths and weaknesses in comparisons. The categorical comparison scores are displayed in Fig. ~\ref{tasks_models}.

\subsection{Analysis of Comparison Results}

The analysis encompasses various aspects of these models, including the number of parameters, response quality, text comprehension capabilities, and language generation abilities.

The examination results have shown that cosmosGPT have achieved quite favorable outcomes in the context of Turkish modeling according to ELO metrics obtained from judge voting. These models have surpassed larger-sized multilingual models while displaying performance that approaches that of large-scale Turkish models.

In categorical tasks, models exhibit strengths over one another in different specific topics. Notably, large models fine-tuned on Turkish have demonstrated significant success in understanding and executing logical and simple mathematical instructions compared to other models. While our smaller parameter cosmosGPT models have struggled in mathematics and logic topics, they are observed to compete at levels comparable to models ten times their size in other subjects.

These findings emphasize that model performance is not solely limited by the number of parameters. cosmosGPT developed for Turkish capture the nuances of the language with fewer parameters and can produce natural, fluent Turkish texts due to the efficiencies in training. A detailed examination of the strengths and weaknesses of language models has shown that cosmosGPT exhibit impressive performance compared to larger models. This underscores the critical role of customized training sets and better language modeling strategies in maximizing the potential of models for languages with limited resources.

\subsection{Inter-Metric Correlation Analysis}

\begin{figure}[htbp]
\centerline{\includegraphics[scale=0.45]{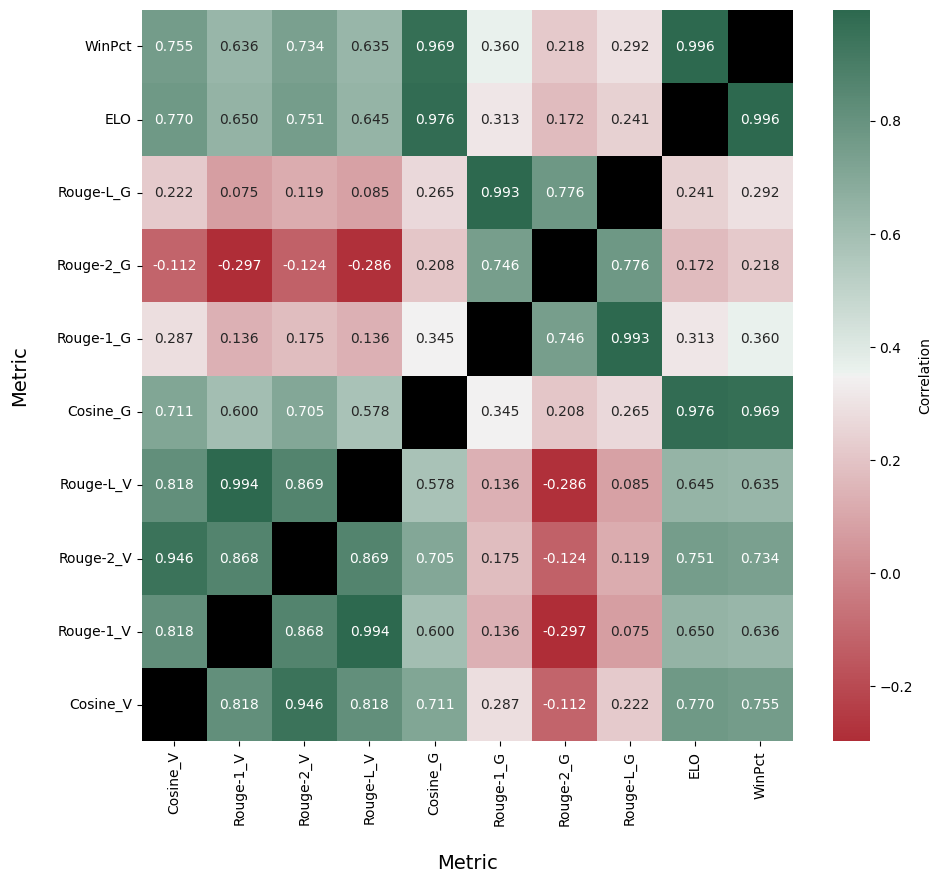}}
\caption{Metric Correlations}
\label{metric_correlations}
\end{figure}

The study has yielded numerous metrics to measure the success of the models. The correlation among these metrics is shown in the Fig. ~\ref{metric_correlations}. It has been observed that the Cosine, ROUGE, ELO, and WinPct metrics derived from the Voting dataset exhibit high correlation with each other. In contrast, the metrics obtained from the General dataset generally show low correlation.

\section{Conclusion and Future Work}

This study has demonstrated that models trained exclusively with Turkish data exhibit promising performance despite their smaller size. While the use of large finetuning datasets, even those with some noise, has proven beneficial, the expected improvements from filtering were not realized. H dataset, entirely collected from human sources, offers significant advantages due to its diverse nature especially when combined with other datasets. Moreover, the capabilities of language models vary significantly from one to another. Our results do not correlate with those from Turkish openLLM\cite{OpenLLMTurkishLeaderboard2024}, highlighting the unique contribution of our study, which measures different capabilities by specifically focusing on human preferences.

Future research will aim to further enhance model performance by incorporating structures such as Direct Preference Optimization (DPO)\cite{rafailov2024direct}, Reinforcement Learning, and others in our fine-tuning processes. By applying these structures, we intend to adapt the models to broader and more diverse datasets, thereby achieving better outcomes. The integration of such advanced methodologies is expected to refine the models’ capabilities, making them more robust and versatile for a variety of linguistic tasks.

Overall, the study underscores the critical role of customized training sets and advanced modeling strategies in maximizing the potential of language models, especially for languages with limited resources. As we continue to refine these models, their ability to understand and produce language will likely improve, providing valuable tools for both academic research and practical applications in NLP.

\section*{Acknowledgment}

Research supported with Cloud TPUs from Google's TPU Research Cloud (TRC).

This study was supported by the Scientific and Technological Research Council of Turkey (TUBITAK) Grant No: 124E055.

\bibliographystyle{ieeetr} 
\bibliography{references} 

\end{document}